\documentclass[runningheads,a4paper]{llncs}

\usepackage{amssymb}
\usepackage{graphicx}
\usepackage{xcolor}
\usepackage{listings}
\usepackage{color}
\usepackage{url}
\usepackage{subfigure}
\usepackage{parcolumns}

\urldef{\mailsa}\path|{gsanthan,|
\urldef{\mailsb}\path|sbasu,|
\urldef{\mailsc}\path|vhonavar}@ist.psu.edu|    
\newcommand{\keywords}[1]{\par\addvspace\baselineskip

\noindent\keywordname\enspace\ignorespaces#1}

\definecolor{gray}{rgb}{0.4,0.4,0.4}
\definecolor{darkblue}{rgb}{0.0,0.0,0.6}
\definecolor{cyan}{rgb}{0.0,0.6,0.6}

\lstdefinelanguage{XML}
{
  morestring=[b]",
  morestring=[s]{>}{<},
  morecomment=[s]{<?}{?>},
  stringstyle=\color{black},
  identifierstyle=\color{cyan},
  keywordstyle=\color{blue},
  morekeywords={xmlns,version,type,VARIABLE,NAME,DOMAIN-VALUE,PREFERENCE-STATEMENT,CONDITION,PREFERENCE,STATEMENT-ID,REGARDLESS-OF}
}

\lstdefinelanguage{SMV}
{
  morestring=[b]",
  morestring=[s]{>}{<},
  morecomment=[s][\itshape\color{gray}]{--}{.},
  commentstyle=\itshape\color{gray},
  stringstyle=\color{black},
  identifierstyle=\color{cyan},
  keywordstyle=\color{blue},
  morekeywords={MODULE,VAR,FROZENVAR,IVAR,INIT,DEFINE,ASSIGN,next,case,esac},
  frame = single, 
  framexleftmargin=1.5pt
}

\begin{document}

\mainmatter  

\title{CRISNER: A Practically Efficient Reasoner for Qualitative Preferences}

\titlerunning{}

%
%
\author{Ganesh Ram Santhanam$^1$ 
\and Samik Basu$^1$ \and Vasant Honavar$^2$}
\authorrunning{Santhanam, Basu \& Honavar}

\institute{$^1$Iowa State University, Ames, Iowa 50011, USA \\
$^2$Pennsylvania State University, University Park, PA 16801, USA\\
$\{$gsanthan,sbasu$\}$@iastate.edu; 
vhonavar@ist.psu.edu}

%
%

\maketitle

\newcommand{\sem}[2]{[\![#1]\!]_{#2}}
\newcommand{\true}{\mathtt{true}}
\newcommand{\false}{\mathtt{false}}
\newcommand{\trans}{\rightarrow}

\newcommand{\ax}[1]{\mathtt{AX}(#1)}
\newcommand{\ex}[1]{\mathtt{EX}(#1)}
\newcommand{\af}[1]{\mathtt{AF}(#1)}
\newcommand{\ef}[1]{\mathtt{EF}(#1)}
\newcommand{\ag}[1]{\mathtt{AG}(#1)}
\newcommand{\eg}[1]{\mathtt{EG}(#1)}
\newcommand{\eu}[2]{\mathtt{E}(#1\ \mathtt{U}\ #2)}
\newcommand{\au}[2]{\mathtt{A}(#1\ \mathtt{U}\ #2)}




\newcommand{\indiff}{\not\succ\!\!\!\not\prec}
\newcommand{\equiva}{\succ\!\!\!\prec}
\newcommand{\R}{{\cal R}\xspace}
\newcommand{\smalltt}[1]{{\mbox{#1}}}
\newcommand{\pref}{\succ}
\newcommand{\satassign}[1]{\mathit{Sat}(#1)}    
\newcommand{\contrib}[1]{\mathit{Contrib}(#1)}
\newcommand{\pset}[1]{\mathcal{P}(#1)}
\newcommand{\prefbrafman}{\ensuremath{\pref^{\circ}}}
\newcommand{\prefwilson}{\ensuremath{\pref^{\!\scriptscriptstyle \blacksquare}}}
\newcommand{\prefb}{\ensuremath{\pref^{\bullet}}}
\newcommand{\precb}{\ensuremath{\prec^{\bullet}}}
\newcommand{\prefm}{\ensuremath{\pref^{m}}}
\newcommand{\precm}{\ensuremath{\prec^{m}}}
\newcommand{\prefa}{\ensuremath{\,\succeq}}
\newcommand{\prefcond}{\ensuremath{\pref^{c}}}
\newcommand{\prefbcond}{\ensuremath{\pref^{\bullet}}}
\newcommand{\precbcond}{\ensuremath{\precb^{c}}}
\newcommand{\prefmcond}{\ensuremath{\prefm}}
\newcommand{\precmcond}{\ensuremath{\precm^{c}}}
\newcommand{\prefacond}{\ensuremath{\prefa}}
\newcommand{\indiffrhd}{\ensuremath{\sim_{\rhd}}}
\newcommand{\rhdclose}{\ensuremath{\rhd}}
\newcommand{\prefp}{\ensuremath{\,{\succ}^{\prime}}}
\newcommand{\prefpeq}{\ensuremath{\,{\succeq}^{\prime}}}

\newcommand{\relimp}{\ensuremath{\rhd}}
\newcommand{\assign}[1]{\ensuremath{\langle {#1} \rangle}}

\newenvironment{mitemize}{
\begin{list}{$\bullet$}{
\setlength{\topsep}{0.1ex} \setlength{\leftmargin}{0.25in}
\setlength{\labelwidth}{1ex} \setlength{\labelsep}{1.0ex}
\setlength{\itemsep}{0.2ex} \setlength{\parsep}{0pt} }}
{\vspace{2pt}\end{list}}

\begin{abstract}
We present CRISNER (Conditional \& Relative Importance Statement Network PrEference Reasoner), a tool that provides practically efficient as well as exact reasoning about qualitative preferences in popular ceteris paribus preference languages such as CP-nets, TCP-nets, CP-theories, etc. The tool uses a model checking engine to translate preference specifications and queries into appropriate Kripke models and verifiable properties over them respectively. The distinguishing features of the tool are: (1) exact and provably correct query answering for testing dominance, consistency with respect to a preference specification, and testing equivalence and subsumption of two sets of preferences; (2) automatic generation of proofs evidencing the correctness of answer produced by CRISNER to any of the above queries; (3) XML inputs and outputs that make it portable and pluggable into other applications. We also describe the extensible architecture of CRISNER, which can be extended to 	new preference formalisms based on ceteris paribus semantics that may be developed in the future.  
\end{abstract}

\section{Introduction}
%
%
Several qualitative preference reasoning languages have been developed in the last two decades, such as CP-nets, TCP-nets, CP-Theories, etc. Despite their relatively high expressive power (compared to quantitative preference formalisms), their widespread application and use in practice is limited, at least in part due to their hardness. The basic reasoning tasks such as dominance and consistency testing are known to be PSPACE-complete for a simple language such as the CP-net. Another reasoning task, checking whether the equivalence or subsumption of the preferences induced by one agent with respect to those induced by another, is important in multi-agent scenarios and negotiation but also known to be hard \cite{Santhanam:ADT2013}. Past works to cope with this hardness include restricting the expressivity of the languages to obtain tractable fragments, and heuristics that yield results in acceptable time but not guaranteed to be exactly correct. Nevertheless, there are applications such as negotiation, planning, security policies, etc. that call for exact reasoning about qualitative preferences with guarantees of correctness (e.g., choosing the best policy to defend a computer network). 

In this paper, we present CRISNER (Conditional \& Relative Importance Statement Network PrEference Reasoner) \cite{CRISNER}, a tool that provides practically efficient as well as exact reasoning about qualitative preferences in popular ceteris paribus\footnote{Ceteris paribus is a Latin term for ``all else being equal''.} \cite{Hansson:JPL1996} preference languages such as CP-nets, TCP-nets and CP-theories. For a preference specification $P$ consisting of a set $\{p_1, p_2, \ldots p_n\}$ of qualitative preference statements in any of the above CP-languages\footnote{Henceforth, we will refer to the languages CP-nets, TCP-nets and CP-theories collectively as `CP-languages' for brevity.}, the ceteris paribus semantics for dominance, consistency etc. are given in terms of reachability over an \emph{induced preference graph} wherein each node corresponds to an outcome and each edge from one node to another represents a preference from the latter node to the former node induced by some statement $p_i$ in $P$. 
CRISNER uses a model checking \cite{Clarke:MIT2000} engine NuSMV \cite{NuSMV} to translate preference specifications and queries into appropriate Kripke structure models  \cite{Kripke:1963} and reachability properties over them respectively. 

\noindent\textbf{Encoding Preferences as Kripke models.\ } Given a specification $P$, CRISNER first succinctly encodes the induced preference graph (IPG) of $P$ into a Kripke structure model $K_P$ in the language of the NuSMV model checker. Although Kripke structures are typically used to represent semantics of temporal and modal logics, we leverage earlier work \cite{Santhanam:AAAI2010,Oster:FACS2012} that demonstrated their use for encoding preference semantics. For testing dominance and consistency with respect to each $P$, CRISNER generates the model $K_P$ only once. Subsequently for each preference query $q$ posed against $P$,  CRISNER translates $q$ into a temporal logic formula $\varphi_q$ in computation-tree temporal logic (CTL) \cite{Queille:SOP1982,Clarke:TOPLAS1986} such that $K_P \models \varphi_q$ if and only if $q$ holds true according to the ceteris paribus semantics of $P$. CRISNER then queries the model checker with the model $K_P$ and  $\varphi_q$ which either affirms $q$ or returns false with a counterexample. For answering queries related to equivalence and subsumption checking of two sets of preferences $P_1$ and $P_2$, CRISNER constructs a combined IPG $K_{P_{12}}$ and uses temporal queries in CTL to identify whether every dominance that holds in $P_1$ also holds in $P_2$ and vice-versa \cite{Santhanam:ADT2013}.

\noindent\textbf{Justification of Query Answers.\ }The answers to queries computed by CRISNER are exact and provably correct for dominance, consistency, equivalence and subsumption queries. Because CRISNER uses the model checker for answering queries, CRISNER is also able to provide proofs or justifications to queries that returned \texttt{false}. CRISNER automatically builds proofs evidencing why the query did not hold true, by collecting and examining the model checker's counterexample and producing a sequence of preference statements whose application proves the correctness of CRISNER's result. 

\noindent\textbf{Tool Architecture.\ }CRISNER is developed in pure Java and is \textit{domain agnostic} in the sense that any set of variables with any domain can be included in a preference specification, although it is optimized for variables with 	binary domains. It accepts preference specifications and queries in a XML format, which provides a common and generic syntax using which users can specify preferences for CP-languages. The results (answers and proofs) for the corresponding queries are also saved in the form XML, so that the results can be further transformed into vocabulary that is more easily understandable by domain users. We describe the architecture of CRISNER and how it can be extended to other ceteris paribus preference formalisms that may be developed in the future. 

CRISNER has been in development for over two years, and to our knowledge, CRISNER is one of the first attempts to develop practical tools for hard qualitative preference reasoning problems. We hope that CRISNER inspires the use of qualitative preference formalisms in practical real world applications, and the development of further qualitative preference languages. 

\vspace{-0.15in}
\section{Background: Syntax and Semantics of CP-languages}
\label{sec:background}
\vspace{-0.1in}
Let $X = \{x_i~|~0 < i \leq n\}$ be a set of preference variables or attributes. For each $x_i \in X$ let $D_i$ be the set of possible values (i.e., domain) such that $x_i = v_i \in D_i$ is a valid assignment to the variable $x_i$. We use $\Phi, \Omega$ (indexed, subscripted or superscripted as necessary) to denote subsets of $X$. The set $\mathcal{O} = \{\alpha~|~\prod_{x_i \in X}{D_i}\}$ of assignments to variables in $X$ is the set of \emph{alternatives}. The \emph{valuation} of an alternative $\alpha \in \cal O$ \emph{with respect to a variable} $x_i \in X$ is denoted by $\alpha(x_i) \in D_i$. 

\vspace{-0.2in}
\subsection{Preference Relations, Statements \& Specifications}
\label{sec:syntax}
\vspace{-0.1in}
Given a set $\mathcal{O}$ of $n$ alternatives, a direct specification of a binary preference relation $\pref$ over $\mathcal{O}$ is difficult, as it requires the user to compare up to $O(n^2)$ pairs of alternatives, which is prohibitive in time. Hence, many preference languages allow for succinct specification of the preference relation over alternatives in terms of \textit{preference relations over the set of attributes that describe the alternatives and their respective valuations or domains}. 

\vspace{-0.1in}
\paragraph{Preference Relations}
Qualitative preference relations can be either (a) \textit{intra-variable preference relations over valuations} of an attribute; or (b) \textit{relative importance preference relations over attributes}. For any $\Phi \subseteq X$, we will use the notation $\pref_{\Phi}$ to denote a preference relation over $D_{\Phi}$, the set of partial assignments to attributes in $\Phi$. For a single attribute $x_i \in X$, the intra-attribute preference relation over its valuations ($D_i$) will be denoted by $\pref_{\{x_i\}}$ or alternatively $\pref_i$ . For example, to formally specify that the valuation $v_i$ is preferred to the valuation $v'_i$ for attribute $x_i$ where $v_i,v'_i \in D_i$, we will write $x_i=v_i \pref_{i} x_i=v'_i$. We will use the notation $\rhd$ to indicate relative importance between attributes or between sets of attributes.

\vspace{-0.1in}
\paragraph{Preference Specifications \& Statements}
In the CP-languages, preferences are expressed in terms of a preference specification, which is a set $P = \{p_i\}$ of preference statements. Each statement $p$ may specify a binary relation over the set $X$ of preference variables (relative importance) and/or a binary relation over the domain of a particular variable (intra-variable preference). The syntax of the preference statements is given below.

\begin{figure}[t]
\begin{center}
    \centering
    \subfigure[CP-net $P_1$]
    {\centering
        \includegraphics[scale=0.37]{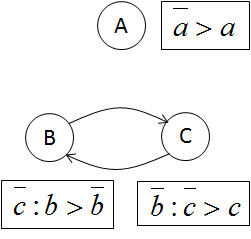}
    }
    \ 
    \centering
    \subfigure[CP-net $P_2$]
    {\centering
        \includegraphics[scale=0.5]{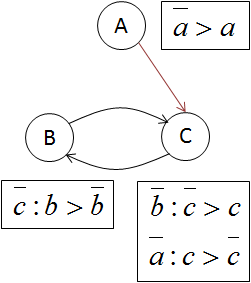}
    }
    \ 
    \centering
	\subfigure[TCP-net $P_3$]
	{
		\centering
		\includegraphics[scale=0.5]{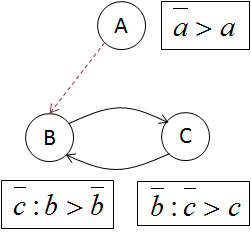}
	}
    \end{center}
        \vspace{-0.15in}
        \caption{Consistent and inconsistent CP-nets; TCP-net\label{fig:cp-net}}
    \vspace{-0.25in}
\end{figure}

CP-nets \cite{Boutilier:JAIR2004,Goldsmith:JAIR2008} allow the specification of only conditional intra-variable preferences\footnote{We use the term CP-nets to refer to the more general formalism defined by Goldsmith et al. \cite{Goldsmith:JAIR2008}.}; TCP-nets \cite{Brafman:JAIR2006} allow the users to specify \emph{pairwise} relative importance among variables in addition to conditional intra-variable preferences as in CP-nets. CP-Theories \cite{Wilson:AI2011} extend TCP-nets by further allowing the specification of the relative importance of one variable over a \emph{set} of variables conditioned on another set of variables. As CP-theories strictly generalize CP-nets and TCP-nets \cite{Wilson:AI2011}, we give the syntax for CP-theories here. A CP-Theory consists of statements of the form 
\vspace{-0.15in}
\begin{center}
$\varrho: x_i=v_i \pref_i x_i=v'_i \ [\Omega]$ 
\end{center}
where $\varrho$ is an assignment to the set $\Phi \subseteq X$ of variables that defines the condition under which the preference holds, $v_i, v'_i \in D_i$, $\Omega \subseteq X$, and $\Phi, \Omega, \{x_i\}$ and $(X - \Phi - \Omega - \{x_i\})$ are disjoint.
The statement expresses the relative importance of the variable $x_i$ over the set $\Omega$ of variables under the condition $\varrho$. Note that CP-nets can be expressed as CP-Theories by fixing $\Omega = \emptyset$ (i.e., $|\Omega| = 0$); and TCP-nets can be expressed as CP-Theories by fixing $|\Omega| = 0$ or $1$. Hence we use the above genera syntactic form to refer to a preference statement in any of the CP-languages. Figure~\ref{fig:cp-net} shows two CP-nets $P_1$ and $P_2$, and a TCP-net $P_3$ where the red dotted arrow from $A$ to $B$ indicates that $A$ is more important than $B$. 
 
\vspace{-0.1in}
\subsection{Ceteris Paribus Semantics}
\label{sec:semantics}
The ceteris paribus semantics of the CP-languages define an induced preference graph with nodes corresponding to the outcomes, and the edges induced by the language-specific interpretation of preference statements. 
According to the \emph{ceteris paribus} interpretation \cite{Hansson:JPL1996,Boutilier:UAI1999},
each preference statement $p \in P$ allows a set of changes to the valuation(s) of one or more variables in an alternative $\beta$ in order to obtain a more preferred alternative $\alpha$, while other variables remain fixed. Such a change is called an \textbf{improving flip}. The improving flips induced by the different types of preference statements is summarized below; 
we do not elaborate on the semantics of each CP-language, and refer the interested reader to \cite{Boutilier:JAIR2004,Brafman:JAIR2006,Wilson:AAAI2004} for details of the semantics of the respective languages.

%

%
%

Consider an intra-variable preference statement $p$ of the form $\varrho: v_i \pref_{\{x_i\}} v'_i$, which can be specified in all the languages we consider. Given two alternatives $\alpha, \beta \in \mathcal{O}$, the ceteris paribus interpretation of the statement $p$ induces an improving flip $(\beta, \alpha) \in E$ in $\delta(P)$ if $\alpha$ and $\beta$ differ only in $x_i$, and $\alpha(x_i)=v_i$ and $\beta(x_i)=v'_i$. In other words for any statement $p$, the valuation of only one variable can be flipped at a time, and that to a more preferred valuation with respect to $p$, while other variables remain fixed. 

\begin{definition}[Preference Semantics for Intra-attribute Preference \cite{Boutilier:JAIR2004}]
\label{def:intravariable-flip}
Given an intra-attribute preference statement $p$ in a preference specification $P$ of the form $\varrho: x_i=v_i \pref_{\{x_i\}} x_i=v'_i$ where $\varrho \in D_{\Phi}, \Phi = \rho(x_i) \subseteq X$ and two alternatives $\alpha, \beta \in \mathcal{O}$, there is an improving flip from $\beta$ to $\alpha$ in $\delta(P)$ induced by $p$ if and only if 
\begin{enumerate}
	\item $\exists x_i \in X : \alpha(x_i)=v_i$ and $\beta(x_i)=v'_i$, 
	\item $\forall x_j \in \Phi : \alpha(x_j) = \beta(x_j) = \varrho(x_j)$, and
	\item $\forall x_k \in X \setminus \{x_i\} \setminus \Phi : \alpha(x_k) = \beta(x_k)$.
\end{enumerate}
\end{definition}

In the above definition, the first condition arises from the intra-attribute preference statement $x_i$; the second condition enforces the condition $\varrho$ in $p$ that states that $\alpha$ and $\beta$ should concur on the parent variables of $x_i$; and the third enforces the ceteris paribus condition that states that $\alpha$ and $\beta$ should concur on all the other variables. 

TCP-nets and CP-Theories allow the specification of relative importance preference of one variable over one or more variable respectively. Hence, multiple variables can change in the same improving flip because a statement of relative importance of one attribute over others means that the user is willing to improve the valuation of the more important attribute \emph{at the expense} of worsening the less important attribute(s). 

\begin{definition}[Preference Semantics for Relative Importance of one Attribute over a Set \cite{Brafman:JAIR2006,Wilson:AAAI2004}]
\label{def:simple-relativeimportance-flip}
Given a relative importance preference statement $p$ in a preference specification $P$ of the form $\varrho: x_i=v_i \pref_{\{x_i\}} x_i=v'_i \ [\Omega]$ where $\varrho \in D_{\Phi}, \Phi \subseteq X$ and two alternatives $\alpha, \beta \in \mathcal{O}$, there is an improving flip from $\beta$ to $\alpha$ in $\delta(P)$ induced by $p$ if and only if 
\begin{enumerate}
	\item $\exists x_i \in X : \alpha(x_i)=v_i$ and $\beta(x_i)=v'_i$, 
	\item $\forall x_j \in \Phi : \alpha(x_j) = \beta(x_j) = \varrho(x_j)$, and
	\item $\forall x_k \in X \setminus \{x_i\} \setminus \Omega \setminus \Phi : \alpha(x_k) = \beta(x_k)$.
\end{enumerate}
\end{definition}

\begin{figure}[t]
\begin{center}
	\centering
    \subfigure[$IPG(P_1)$]
    {\centering
        \includegraphics[scale=0.4]{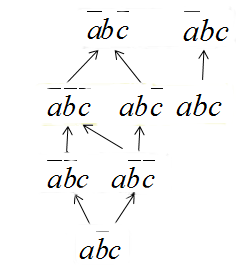}
    }
    \ 
	\centering
    \subfigure[$IPG(P_2)$]
    {\centering
        \includegraphics[scale=0.5]{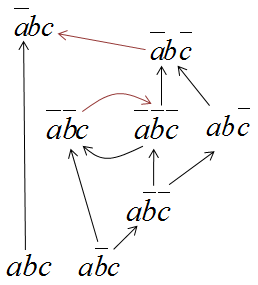}
    }
    \ 
	\centering
	\subfigure[$IPG(P_3)$]
	{
		\centering
		\includegraphics[scale=0.4]{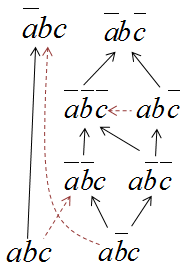}
	}
	
\end{center}
    \vspace{-0.2in}
\caption{Induced preference graphs of CP-nets and TCP-net \label{fig:ipg}}
    \vspace{-0.2in}
\end{figure}

In the above definition, the first condition arises from the preference statement $p$ on $x_i$; the second condition enforces the condition $\varrho$ in $p$; and the third enforces the ceteris paribus condition and allows for unrestricted changes to the attributes that are less important than $x_i$ (in $\Omega$) when this preference statement is applied. 

The above interpretations define valid improving flips induced by statements in $P$ that correspond to edges in the \textbf{induced preference graph} denoted $IPG(P)$, whose nodes are the set of all outcomes or alternatives (i.e., the set of all assignments to all preference variables), and which represents the dominance relation $\pref_P$ over the outcomes. Figure~\ref{fig:ipg} shows the induced preferences graphs for the respective CP-nets and TCP-net in Figure~\ref{fig:cp-net}. Note that the red solid edges in $IPG(P_2)$ are those induced by the dependency of $C$ on $A$ in $P_2$. Similarly, the red dotted edges in $IPG(P_3)$ are those induced by the relative importance of $A$ over $B$ in $P_3$.

%

\vspace{-0.2in}
\section{Preference Queries}
\label{sec:queries}
\vspace{-0.15in}
Computing answers to preference queries with respect to a given preference specification in the ceteris paribus semantics amounts to making querying properties related to reachability on the induced preference graph. We consider the following preference queries that have been implemented in CRISNER in this paper. 

\begin{definition}[Dominance \& Consistency Testing \cite{Boutilier:JAIR2004,Goldsmith:JAIR2008,Wilson:AI2011}]
Given a preference specification $P$ consisting of a set of preference statements $\{p_1, p_2 \ldots p_n\}$, and two outcomes $\alpha, \beta \in \mathcal{O}$  
\begin{enumerate}
	\item \textbf{Dominance Testing ($\alpha \pref_P \beta$)}  asks whether there a sequence of improving flips from $\beta$ to $\alpha$ in $IPG(P)$?
	\item \textbf{Consistency Testing} asks whether the preferences induced by $P$ are consistent, i.e., is there is a cycle in $IPG(P)$? 
\end{enumerate}
\end{definition}

\begin{definition}[Preference Equivalence \& Subsumption \cite{Santhanam:ADT2013}]
Given two preference specifications $P_1, P_2$ and two outcomes $\alpha, \beta \in \mathcal{O}$,
\begin{enumerate}
	\item \textbf{Preference Subsumption ($P_1 \sqsubseteq P_2$)} asks whether $\alpha \pref_{P_1} \beta \Rightarrow \alpha \pref_{P_2} \beta$.
	\item \textbf{Preference Equivalence ($P_1 \equiv P_2$)} asks whether $P_1 \sqsubseteq P_2$ and $P_2 \sqsubseteq P_1$.
\end{enumerate}
\end{definition}


\subsection{Preference Query Answering via Model Checking}
\label{sec:reasoning-via-mc}
In order to address the PSPACE-hardness of the problems of dominance and consistency, CRISNER implements the model checking based approach to preference query reasoning as presented in the series of works by Santhanam et al. \cite{Santhanam:AAAI2010,Santhanam:JAIR2011,Santhanam:ADT2013}. There are two direct benefits of using the model checking approach to answering preference queries. First, by using the model encoding techniques presented in the above works, then dominance and consistency queries can be transformed into equivalent reachability queries on the Kripke model in a straightforward way. The transformation from the preference specification and query to the Kripke model and temporal logic formula respectively is at a syntactic level prior to execution of the model checker (without having to build the induced preference graph); thus preserving the benefits of succinctness of the CP-languages. When the query is actually executed on the model checker, the induced preference graph is expanded to the extent needed by the model checker to verify the temporal logic formula. This would enable us to leverage the decades of advances in model checking algorithms and tools. The second direct benefit if the model checker returns a counter example to the encoded temporal logic formula (corresponding to the original preference query), which is in terms of states and transition sequences in the Kripke model, then it can be conveniently mapped back to nodes and paths in the induced preference graph. This allows CRISNER to automatically generate justifications for dominance and consistency queries by producing sequences of improving flips in the induced preference graph that either prove the dominance or disprove the consistency. 

\paragraph{CTL model checking}
We use formulas in computation-tree temporal logic (CTL) \cite{Clarke:MIT2000} for verifying reachability within the Kripke model generated by CRISNER. Our choice of model checker is NuSMV \cite{NuSMV}, an open source and widely used tool. CTL is an extension of propositional logic; CTL uses propositional and temporal connectives to express temporal properties, whose semantics is given in terms of a set of states in the Kripke structure where the properties are satisfied. We briefly outline the syntax and semantics of some CTL connectives below, and refer the reader to \cite{Clarke:MIT2000} for details. The syntax of CTL is described as follows:
\[
\varphi \rightarrow
true~|~\texttt{Atomic-Propositions}~|~\neg\varphi~|~\varphi\lor\varphi~|~\texttt{EX}\varphi~|~\texttt{EG}\varphi~|~\texttt{E}(\varphi\ \texttt{U}\ \varphi)
\]
The semantics of a CTL formula $\varphi$ is given in terms of the set of states in a Kripke structure that satisfy $\varphi$. The
propositional constant $true$ is satisfied in all states; the
proposition $p$ holds in states labeled with $p$. The negation of $\varphi$ is satisfied if the formula does not hold. The disjunct of two CTL formulas is satisfied by states if at least one of the disjuncts is satisfied. The rest of the operators in the CTL syntax are temporal operators that quantify
the states and the transitions. The property $\texttt{EX}\varphi$ is
satisfied in any state $s$ which can reach some (\texttt{E}, for existential quantification) state $t$ in one transition (\texttt{X}, for one step reachability) such that $t$ satisfies $\varphi$. The property $\texttt{EG}\varphi$ is satisfied in any state $s$ which has
some (\texttt{E}) path where every (\texttt{G}) state $t_i$'s in the
path satisfy $\varphi$. Finally, the property $\texttt{E}(\varphi_1\ \texttt{U}\ \varphi_2)$ is satisfied in any state
$s$ which has some (\texttt{E}) path where there exists a state $t$
which satisfied $\varphi_2$ and in all states before $t$, $\varphi_1$
is satisfied. Semantics for other CTL formulas are described in terms of the above, e.g., $\texttt{EF}\varphi$ is satisfied in any state from where
there exists a path eventually leading to a state that satisfies
$\varphi$. This is equivalent to $\texttt{E}(true\ \texttt{U}\ \varphi)$. There are other temporal connectives which we mention and explain as needed in terms of the above in this paper.
	
\subsection{Encoding Preference Queries in CTL}
\label{sec:pref-queries}
Here we outline the overall strategy to obtain answers to preference queries in the languages CP-languages. We assume that CRISNER is given a preference specification $P$ and a preference query $q$ that is a test for either a dominance, consistency, preference subsumption or preference equivalence. The task of CRISNER is then to compute whether $q$ holds (or not) with respect to the ceteris paribus semantics of $P$.

\paragraph{Dominance Testing}
Testing dominance of an outcome over another amounts to simply checking reachability from one outcome to another in $IPG(P)$. To answer the dominance query $\alpha \pref_P \beta$, CRISNER constructs a temporal logic formula that corresponds to reachability from $\beta$ to $\alpha$ in $IPG(P)$ and executes the query on the NuSMV \cite{NuSMV} model checker, affirming dominance if and only if the model checker returns true. Defining $AP$ to be the set of preference variables in $P$ and $\varphi_{\alpha}$ to be the formula encoding the set of variables assigned $true$ in the outcome $\alpha$, the above dominance can be encoded in the following CTL formula: 
\vspace{-0.05in}
\begin{center}
$\varphi_{\beta} \Rightarrow \texttt{EF} \varphi_{\alpha}$
\end{center}

\vspace{-0.15in}
\paragraph{Consistency Testing}
Consistency testing amounts to checking that there are no cycles in $IPG(P)$. To test consistency for $P$, CRISNER verifies the following CTL formula, which states that for all \texttt{start} states  (If no start states are initialized, NuSMV non-deterministically considers all states in the Kripke model as start states	), there must be no reachable node from which there is a path back to $\alpha$, i.e. there is no cycle in $IPG(P)$. 
\vspace{-0.1in}
\begin{center}
$\mbox{start} \Rightarrow \neg \texttt{EX}\ (\mbox{g}=1 \land \texttt{EF}\ \mbox{start})$
\end{center}
In the above, $g=1$ holds whenever the destination state of a current state results from an improving flip as per the underlying preference semantics.
\vspace{-0.15in}
\paragraph{Preference Subsumption and Equivalence}
Preference equivalence (subsumption) checking \cite{Santhanam:ADT2013} amounts to testing whether the preferences induced by two preference specifications $P_1$ and $P_2$ are the same (or such that one subsumes the other) or not. Given $P_1$ and $P_2$, CRISNER constructs an SMV model $K_{P_{12}}$ of the \emph{combined induced preference graph} (CIPG) \cite{Santhanam:ADT2013} that encodes the union of preferences induced by $P_1$'s preference statements and the reverse of those induced by $P_2$'s preference statements. In order to answer whether $P_1$ subsumes the preferences induced by $P_2$ or not, CRISNER constructs the following temporal logic formula that is verified by NuSMV if and only if $P_1$ subsumes $P_2$ (we use $g_1$ and $g_2$ to refer to the $g$ for the models corresponding to $P_1$ and $P_2$ respectively. 
\vspace{-0.1in}
\begin{center}
$\varphi : \texttt{\textbf{AX}} \ \big( \ g_1 \Rightarrow \texttt{\textbf{EX}} \ \ \texttt{\textbf{E}} \ \big[ g_2 \ \texttt{\textbf{U}} \ \ (\mbox{start} \ \land \  g_2) \ \big] \ \big)$
\end{center}
The above states that whenever there is an improving flip in $IPG(P_1)$ from $\alpha$ to an outcome ($\texttt{AX} g_1$..), then there exists a sequence of improving flips ($\texttt{EX} E(g_2 \texttt{U} ...)$) from that outcome in $IPG(P_2)$ back to $\alpha$. In the above, all possible $\alpha$ is captured by the proposition $\texttt{start}$.
Preference equivalence is checked by in turn verifying that $P_1 \sqsubseteq P_2$ and $P_2 \sqsubseteq P_1$.

\section{XML Input Language}
\label{sec:xml-input}
\vspace{-0.1in}
CRISNER accepts a preference specification for any of the CP-languages in an XML format. The preference specification consists of a declaration of the preference variables, their domains and a set of preference statements. Each preference statement is of the form discussed in Section~\ref{sec:syntax}, and expresses an intra-variable and/or relative importance preference relation over the domain of a variable. 
\vspace{-0.15in}
\subsection{Defining Preference Variables}
\vspace{-0.1in}
Figure~\ref{fig:var-def-xml} shows part of a preference specification defining variables and their domains. The preference variable $a$ has a binary domain with values $0$ and $1$, whereas $x$ has a domain $\{0, 1, 2\}$. Note that CRISNER supports domain valuations with string values that are combinations of letters and numbers, as allowed by NuSMV.
\begin{figure}[h]
\noindent\begin{minipage}{.45\textwidth}
\footnotesize
\vspace{-0.3in}
\lstset{language=XML,alsodigit={-}}
\begin{lstlisting}
<VARIABLE>
  <NAME>a</NAME>
  <DOMAIN-VALUE>0</DOMAIN-VALUE>
  <DOMAIN-VALUE>1</DOMAIN-VALUE>
</VARIABLE>
\end{lstlisting}
\centering{(a)}
\end{minipage}\hfill
\begin{minipage}{.45\textwidth}
\vspace{-0.3in}
\lstset{language=XML,alsodigit={-}}

\begin{lstlisting}
<VARIABLE>
  <NAME>x</NAME>
  <DOMAIN-VALUE>0</DOMAIN-VALUE>
  <DOMAIN-VALUE>1</DOMAIN-VALUE>
  <DOMAIN-VALUE>2</DOMAIN-VALUE>
</VARIABLE>
\end{lstlisting}
\centering{(b)}
\end{minipage}\hfill
\vspace{-0.05in}
\caption{XML encoding of definitions of preference variable $a$ with domain size 2 and 3}
\label{fig:var-def-xml}
\vspace{-0.4in}
\end{figure} 

\subsection{Specifying Conditional Preference Statements}
\vspace{-0.1in}
The listing in Figure~\ref{fig:cp-net-xml} shows a portion of a preference specification that declares preferences over values of the variable $c$ conditioned on the variables $b$ and $a$ respectively. The $\texttt{VARIABLE}$ tag identifies the variable whose preferences are being specified. Note that there can be multiple conditions or no conditions as well. In addition, there can also be multiple preferences for a variable, e.g., if there is a variable with domain of ${0, 1, 2}$ then to specify the total order $0 \pref 1 \pref 2$ one would encode $0 \pref 1$ as one preference followed by $1 \pref 2$. CRISNER requires that the variable names and their assignments match with the preference variable declarations in the file; otherwise the tool reports an appropriate error stating that the variable is not defined in the preference specification.

\begin{figure}[t]
\noindent\begin{minipage}{.45\textwidth}
\footnotesize

\lstset{language=XML,alsodigit={-}}
\begin{lstlisting}
<PREFERENCE-STATEMENT>
  <STATEMENT-ID>p3</STATEMENT-ID>
  <VARIABLE>c</VARIABLE>
  <CONDITION>b=0</CONDITION>
  <PREFERENCE>0:1</PREFERENCE>
</PREFERENCE-STATEMENT>
\end{lstlisting}
\centering{(a)}
\end{minipage}\hfill
\begin{minipage}{.45\textwidth}

\lstset{language=XML,alsodigit={-}}
\begin{lstlisting}
<PREFERENCE-STATEMENT>
  <STATEMENT-ID>p4</STATEMENT-ID>
  <VARIABLE>c</VARIABLE>
  <CONDITION>a=0</CONDITION>
  <PREFERENCE>1:0</PREFERENCE>
</PREFERENCE-STATEMENT>
\end{lstlisting}
\centering{(b)}
\end{minipage}\hfill
\caption{XML encoding of conditional preference statements $p_3$ and $p_4$ in a CP-net}
\label{fig:cp-net-xml}
\end{figure} 

\vspace{-0.15in}
\subsection{Specifying Relative Importance Preferences}
\vspace{-0.1in}
In order to allow specification of relative importance of one variable over another, as in a TCP-net, CRISNER allows the tag $\texttt{REGARDLESS-OF}$ within a preference statement. Figure~\ref{fig:tcp-net-cpt-xml}(a) declares a preference statement that says (in addition to the fact that $a=0 \pref_a a=1$)  that $a$ is relatively more important than $b$. In order to specify relative importance of one variable over a set of other variables (simultaneously) as allowed by a CP-theory, the user can specify multiple $\texttt{REGARDLESS-OF}$ tags within the same preference statement. For instance, Figure~\ref{fig:tcp-net-cpt-xml}(b) shows a preference statement that declares that $a$ is relatively more important than $\{b,c\}$. 
\begin{figure}[h]
\noindent\begin{minipage}{.45\textwidth}
\footnotesize
\vspace{-0.1in}
\lstset{language=XML,alsodigit={-}}
\begin{lstlisting}
<PREFERENCE-STATEMENT>
  <STATEMENT-ID>p1</STATEMENT-ID>
  <VARIABLE>a</VARIABLE>
  <PREFERENCE>0:1</PREFERENCE>
  <REGARDLESS-OF>b</REGARDLESS-OF>
</PREFERENCE-STATEMENT>
\end{lstlisting}
\centering{(a)}
\end{minipage}\hfill
\begin{minipage}{.45\textwidth}
\vspace{-0.15in}
\lstset{language=XML,alsodigit={-}}
\begin{lstlisting}
<PREFERENCE-STATEMENT>
  <STATEMENT-ID>p1</STATEMENT-ID>
  <VARIABLE>a</VARIABLE>
  <PREFERENCE>0:1</PREFERENCE>
  <REGARDLESS-OF>b</REGARDLESS-OF>
  <REGARDLESS-OF>c</REGARDLESS-OF>
</PREFERENCE-STATEMENT>
\end{lstlisting}
\centering{(b)}
\end{minipage}\hfill
\caption{XML encoding of conditional preference statements $p_3$ and $p_4$ in a CP-net}
\label{fig:tcp-net-cpt-xml}
\vspace{-0.15in}
\end{figure} 

\section{Encoding Preferences as SMV Models}
\vspace{-0.15in}
CRISNER encodes Kripke models for a preference specification as described earlier. Now we discuss constructs in NuSMV used by CRISNER for the encoding. 
\vspace{-0.35in}
\subsection{Encoding Preference Variables \& Auxiliary Variables}
\vspace{-0.1in}
In order to encode the CP-net $P_1$ in our earlier example, CRISNER generates the code for the SMV model as shown in Figure~\ref{fig:smv-cp-net}. We explain the translation of a preference specification into an SMV model by CRISNER through this example.

CRISNER defines just the $\texttt{main}$ module, with 3 variables corresponding to the preference variables in $P_1$ and another variable $g$, which we will explain shortly. 
We overload $a, b, c$ to refer to variables in the Kripke model and variables in the preference specification, hence valuations of $a, b, c$ in a state $s$ in the Kripke model respectively correspond to the valuations of the preference variables $a, b, c$ in the preference specification $P$. The variables $a, b, c$ are \emph{state} variables in the SMV model i.e., their valuations stored by the model checker for each state explored during model checking. 

The \texttt{IVAR} variables $cha, chb, chc$ are modeled as \emph{input} variables, i.e., their valuations are not stored as part of each state. The model checker initializes them non-deterministically for each state so that all paths are open for exploration by the model checker during verification.  Each preference statement is translated into an appropriate guard condition for a transition in the Kripke model, and the semantics of variables $cha, chb, chc$ either allows or disallows the change in the value of the corresponding preference variable $a, b, c$, in accordance with the improving flip semantics.   

\noindent\textbf{Identifying transitions corresponding to improving flips.\ }
The additional $g$ variable is defined to be $true$ exactly when the model checker transitions from a state corresponding to one outcome to a state corresponding to another outcome (not transitions between states corresponding to the same outcome). Hence, we can conveniently refer to transitions in the Kripke models that correspond to improving flips by constraining $g$ to have valuation $1$ in the destination state.

\noindent\textbf{Referencing start states explored by NuSMV.\ } The \texttt{FROZEN} variables \texttt{a\_0, b\_0, c\_0} are constrained to be fixed with the values of the variables $a, b, c$ respectively at the start of the model checking algorithm via the \texttt{DEFINE} and \texttt{INIT}  constructs. This allows us to refer to the state non-deterministically picked by the model checker as the start of model exploration using $start$. This is used for computing consistency, preference subsumption and preference equivalence.

\begin{figure}[h]
\noindent\begin{minipage}{.34\textwidth}
\footnotesize
\vspace{-0.3in}
\lstset{language=SMV,alsodigit={-}}
\begin{lstlisting}
MODULE main
VAR
  a : {0,1};
  c : {0,1};
  b : {0,1};
  g : {0,1};
FROZENVAR
  a_0 : {0,1};
  b_0 : {0,1};
  c_0 : {0,1};
IVAR
  cha : {0,1};
  chb : {0,1};
  chc : {0,1};
DEFINE
start := a=a_0 
    & b=b_0 & c=c_0;
    
INIT start=TRUE;

\end{lstlisting}
\end{minipage}\hfill
\begin{minipage}{.6\textwidth}
\vspace{-0.3in}
\lstset{language=SMV,alsodigit={-}}
\begin{lstlisting}
ASSIGN
 next(a) := case
  a=1 & cha=1 & chb=0 & chc=0 : 0;
  TRUE : a;
 esac;
 next(c) := case
  c=1 & b=0 & cha=0 & chb=0 & chc=1 : 0;
  TRUE : c;
 esac;
 next(b) := case
  b=0 & c=0 & cha=0 & chb=1 & chc=0 : 1;
  TRUE : b;
 esac;
 next(g) := case
  a=1 & cha=1 & chb=0 & chc=0 : 1;
  c=1 & b=0 & cha=0 & chb=0 & chc=1 : 1;
  b=0 & c=0 & cha=0 & chb=1 & chc=0 : 1;
  TRUE: 0;
 esac;
\end{lstlisting}
\end{minipage}\hfill
    \vspace{-0.15in}
\caption{SMV code for Kripke Model encoding $IPG(P_1)$}
    \vspace{-0.25in}
\label{fig:smv-cp-net}
\vspace{-0.25in}
\end{figure}
%
%
%
\subsection{Encoding Preference Statements}
\vspace{-0.05in}
\noindent\textbf{Encoding Intra-variable Preferences.\ } 
To encode a intra-variable preference statement for a variable $x$ with a condition $\rho$ on the other variables, the \texttt{next(x)} construct encodes guards such that the valuations of the other variables correspond to those in the condition $\rho$, and valuation of $chx$ is $1$ while all other $ch$ variables are set to $0$. As an example, \texttt{next(a)} includes a transition such that $c$ changes from $1$ to $0$ precisely when $b=0$ and $chc=1$ ($cha=0, chb=0$, allowing only $c$ to change in that transition), which corresponds to the improving flip induced by $p_3$ conditioned on the value of $a$ in the CP-net $P_1$ (Figure~\ref{fig:smv-cp-net}). 

\noindent\textbf{Encoding Relative Importance.\ }
For modeling relative importance preference statements, multiple \texttt{IVAR} variables can be assigned $1$ in guard conditions such that the more important and less important preference variables can change in the same transition - corresponding to an improving flip for relative importance. For example, Figure~\ref{fig:smv-relimp} shows a snippet of the SMV code that models the transitions arising from the relative importance of $a$ over $b$ as in the TCP-net $P_3$ shown in Figure~\ref{fig:cp-net}(c). Note that $cha$ and $chb$ are set to $1$ for the second guard condition of \texttt{next(b)}, allowing $a$ to change to a preferred value trading off $b$. In order to model relative importance as in a CP-theory where one variable is more important than multiple others, a similar encoding is used, except that all the corresponding $ch$ variables are set to $1$. 
\begin{figure}[h]
\begin{center}
\vspace{-0.15in}
\lstset{language=SMV,alsodigit={-}}
\begin{lstlisting}
  next(b) :=  case
      b=0 & c=0 & cha=0 & chb=1 & chc=0 : 1; 
      a=1 & cha=1 & chb=1 & chc=0 : {0,1};
      TRUE : b;
  esac;
\end{lstlisting}
    \vspace{-0.2in}
\caption{Encoding relative importance preferences for the TCP-net $P_3$}
    \vspace{-0.1in}
\label{fig:smv-relimp}

\end{center}
\vspace{-0.2in}
\end{figure} 

\subsection{Justification of Query Results}
    \vspace{-0.05in}
In addition to answering preference queries posed against preference specifications, CRISNER also provides a justification of the result where appropriate. 
In order to obtain justification, CRISNER uses the counterexamples returned by NuSMV model checker whenever a temporal logic formula is not satisfied. 

\medskip
\noindent\textbf{Extracting a Proof of Dominance.\ }
In the case of a dominance query, if CRISNER returns \texttt{true}, we construct another temporal logic formula that states the negation of the dominance relationship, which obtains a sequence of outcomes corresponding to an improving flipping sequence from the lesser preferred to the more preferred outcome from the model checker. Suppose that we want to obtain proof that an alternative $\alpha$ dominates another alternative $\beta$. This means that $\varphi_{\beta} \rightarrow \texttt{EF} \varphi_{\alpha}$ holds. We then verify $\neg(\varphi_{\beta} \rightarrow \texttt{EF} \varphi_{\alpha})$, which obtains a sequence of states in the Kripke model corresponding to an improving flipping sequence from $\beta$ to $\alpha$ from the model checker corresponding to an improving flipping sequence from $\beta$ to $\alpha$ which serves as the proof of dominance. 

\medskip
\noindent\textbf{Extracting a Proof of Inconsistency.\ }
In the case of a consistency query (see Section~\ref{sec:pref-queries}), CRISNER returns a sequence of outcomes corresponding to an improving flipping sequence from an outcome to itself (indicating a cycle in the induced preference graph) whenever the preference specification input is inconsistent. 

\medskip
\noindent\textbf{Extracting a Proof of Non-subsumption.\ } For a preference subsumption query $P_1 \sqsubseteq P_2$, CRISNER provides an improving flip from one outcome to another induced by $P_1$ but not induced by $P_2$ whenever the query does not hold.

In the above, counterexamples returned by NuSMV are in terms of states and transitions in the Kripke model; CRISNER parses and transforms the counterexamples back into a form that relates to the preference variables, outcomes and improving flips in the induced preference graph of the preference specification, and saves it in an XML format. 

    \vspace{-0.2in}
\section{Architecture}
\begin{figure}[t]
\includegraphics[scale=0.48]{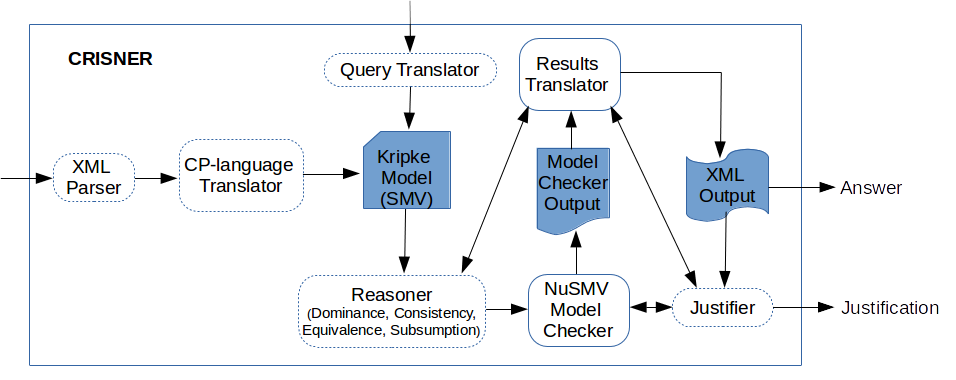}
\label{fig:architecture}
    \vspace{-0.35in}
\caption{Architecture and components of CRISNER preference reasoner}
\vspace{-0.25in}
\end{figure}

    \vspace{-0.1in}
CRISNER is built using the Java programming language\footnote{Third party libraries used by CRISNER are listed in the project site \cite{CRISNER}.}. The architecture of CRISNER consists of several components as depicted in Figure~\ref{fig:architecture}. 

The XML parser is used to parse the preference specifications and preference queries input by the user\footnote{While currently CRISNER does not use XML schema or DTD to validate the XML input, we plan to enforce that in future.}. 
The CP-language translator is a critical component that constructs the SMV code for the Kripke model for the preference specification input. It declares the necessary variables with their domains, sets up the \texttt{DEFINE}, \texttt{TRANS} and \texttt{INIT} constraints and finally generates guard conditions for enabling transitions corresponding to improving flips induced by each preference statement (as discussed in Section~\ref{sec:reasoning-via-mc}). 

CRISNER provides two interfaces for preference reasoning. The first is a simple command line menu-drive console interface where the user can provide either (a) one preference specification as input and then use the menu to pose dominance and consistency queries, or (b) two preference specifications as input and pose a subsumption or equivalence query. The answer (\texttt{true/false}) obtained and the justification for the answer (where possible) is provided on the console. 
Another way of using CRISNER is to specify preference queries in an XML file that contains a dominance or consistency or preference equivalence or subsumption query, and identifies the preference specification against which the query should be executed. 
The query translator component parses queries specified in XML format and provides it to the Reasoner component.
Further details about the XML tags and examples of preference specification, preference queries and sample SMV code generated by CRISNER are available from CRISNER's project website \url{http://www.ece.iastate.edu/~gsanthan/crisner.html}.

The Reasoner is another critical component in CRISNER that constructs a temporal logic formula corresponding to the preference query posed by the user, and invokes the NuSMV model checker to verify the formula. The result and any counter examples generated by the model checker are parsed by the Results Translator, and saved in XML format by the XML Encoder. If a counter example is applicable to the preference query, then the Justifier parses the XML output and executes any followup queries on the model checker (e.g., verification of the negation of the dominance query) to provide the user with the appropriate proof. 

    \vspace{-0.1in}
\paragraph{Extensibility.\ }
Although CRISNER currently supports the CP-net, TCP-net and CP-theory formalisms, it can be extended to support another qualitative preference language, as long as the semantics of the language is described in terms of an induced preference graph. 
To extend support, the XML parser must be extended to support the syntax of the new language; and the CP-language translator must be extended to generate SMV code according to the semantics of the new language. 
We are currently working on including support for the conditional importance network (CI-net) \cite{Bouveret:IJCAI2009} language for representing and reasoning with preferences in the context of fair division of indivisible goods.
As another example, one can think of a new preference formalism may allow expression of preferences of one partial assignment of variables over another. In such a case, the CP-translator component can be extended to include a translation for such a preference statement into guard conditions in the Kripke model representing the induced preference graph. 
CRISNER can also be extended to support new preference queries, for example, computing a weak order or total order extension of the partial order induced by a preference specification. In this case the Query translator and Reasoner should be extended to translate the new queries into one or more appropriate temporal logic formulas (as described in \cite{Oster:FACS2012}) and the Justifier should be extended to construct and execute  follow up queries that obtain proofs for the answers.
    \vspace{-0.1in}
\paragraph{Scalability.\ }
While we have not yet performed a systematic experiment studying CRISNER's runtime performance for preference specifications of different sizes (number of preference variables), our preliminary tests have revealed that CRISNER answers dominance (including the computation of justification for a when applicable) in less than a minute on average for upto 30 variables on a 8GB Corei7 Windows 7 desktop. 
Although CRISNER allows variables with domain size $n > 2$, the model checking performance degrades quickly with increasing $n$; this can be alleviated by configuring the NuSMV model checker to use multi-way decision diagrams \cite{NuMDG}.

\vspace{-0.2in}
\section{Concluding Remarks}
    \vspace{-0.1in}
We presented CRISNER, a tool for specifying and reasoning with qualitative preference languages such as CP-net, TCP-net and CP-theory. CRISNER translates preference specifications and queries with respect to them provided in XML format into a Kripke structure and corresponding temporal logic (CTL) queries in the input language of the NuSMV model checker. Currently CRISNER supports dominance, consistency, preference equivalence and subsumption testing for the above languages. The obtained results from the model checker, including proofs of dominance, inconsistency or non-subsumption, are translated by CRISNER back in terms of the vocabulary of the input preference specification and saved in XML format. CRISNER's architecture supports extension to other preference queries and preference languages such as CI-nets whose semantics are in terms of the induced preference graph.

Work in progress includes adding support for dominance and consistency testing with CI-nets, which is useful in multi-agent fair division problems. In the future, we plan to add support for computing weak order and total order extensions to consistent and inconsistent preference specifications, which is useful in applications like stable stable matching and recommender systems. We also plan to add support for reasoning with group preferences, i.e., reasoning about the preferences of multiple agents. 

\vspace{-0.1in}
\bibliographystyle{splncs03}

\bibliography{references,references-tocl,ganesh}

\end{document}